# Towards explainable classifiers using the counterfactual approach - global explanations for discovering bias in data

Agnieszka Mikołajczyk
Department of Electrical Engineering, Control Systems and
Informatics, Gdansk University of Technology, Poland

Michał Grochowski
Department of Electrical Engineering, Control Systems and
Informatics, Gdansk University of Technology, Poland

Arkadiusz Kwasigroch
Department of Electrical Engineering, Control Systems and
Informatics, Gdansk University of Technology, Poland

May 12, 2020

**Abstract** The paper proposes summarized attribution-based post-hoc explanations for the detection and identification of bias in data. A global explanation is proposed, and a step-by-step framework on how to detect and test bias is introduced. Since removing unwanted bias is often a complicated and tremendous task, it is automatically inserted, instead. Then, the bias is evaluated with the proposed counterfactual approach. The obtained results are validated on a sample skin lesion dataset. Using the proposed method, a number of possible bias-causing artifacts are successfully identified and confirmed in dermoscopy images. In particular, it is confirmed that black frames have a strong influence on Convolutional Neural Network's prediction: 22% of them changed the prediction from benign to malignant.

## 1 Introduction

In recent years, deep neural networks (DNNs) have achieved state-of-the-art performance in various tasks. Currently, in contrast to shallow models exploited in the past, most of deep systems extract features automatically, and to do that, they tend to rely on a vast number of labeled data. Whereas the quality of datasets used to train neural networks has a significant impact on model's performance, those datasets are often noisy, biased, and sometimes even contain incorrectly labeled samples [1]. Moreover, DNNs usually have tens of layers, with millions of parameters and very complex latent space, which makes them very hard to interpret.

Nevertheless, those fragile black-box deep machine learning models are used to solve sensitive and critical tasks, where the demand for clear reasoning and correct decision is high [2]–[4]. Hence, there is raising awareness towards robust learning, formal verification, and extensive testing of models. However, without knowing that data is biased, training the model is a tricky and challenging task.

The paper proposes a method to detect bias in data with attribution-based locally-summarized global explanations, coming from post-hoc Explainable Artificial Intelligence (XAI). This method is given the name of **GEBI** – **G**lobal **E**xplanations for **B**ias **I**dentification. Focus is put on image classification and testing it on the skin lesion recognition task, however, GEBI can be applied to any other problem as well.

The proposed global explanation method is an improvement of the first global analyzer dedicated to summarizing attribution-based explanations automatically (Spectral

Relevance Analysis - SpRAy [5]). The newly proposed solution aims to compensate the previously unnoticed drawback of biased XAI, which strongly focused on localization and shape of model's attribution but completely ignored an essential part of the explanation: *why it focuses there*? The improved algorithm of summarized global, relevance-based, post-hoc explanations for discovering biases in data takes inspiration from how humans analyze visual explanations: an attribution map and input image altogether. In particular, the paper describes a novel GEBI method of global post-hoc explainability to help explain deep neural network decisions, to justify them, to control their reasoning process, and to discover new knowledge. Moreover, a simple framework is proposed on how to measure the impact of possible bias-causing artifacts with a counterfactual approach. The counterfactual analysis evaluates how the change of input features changes the predicted output [6]. Since removing the unwanted bias is often a complicated and tremendous task, it is automatically inserted, instead. The process of bias insertion helps a user to understand the causes of model's decision making [7].

Then, the effect of insertion of such bias on the prediction change is measured.

The major contribution of the paper includes:
- proposition of a GEBI method to improve SpRAy by analyzing the explanation (attribution map) along with the input,
- proposition of a counterfactual approach for bias testing with the proposed bias insertion algorithm.

In the Related works section, the subject of explainable artificial intelligence is brought closer, along with a brief review of what approaches have been made in the past to uncover biases in data collections. Then, the next section gives a detailed methodology description. In the Experiments section, the operation of the proposed algorithm is demonstrated on the example of a skin lesion dataset. The detected clusters are manually examined and analyzed to find prediction patterns. Then, after detecting artifacts that might cause bias, the nature and scale of prediction changes caused by the presence of such artifact is measured. Finally, the discussion of the obtained results is presented, along with the proposal on how to improve the biased model.

## 2 Related works

In this section, the subject of explainable artificial intelligence is brought closer, along with a brief review on what approaches have been made in the past to uncover bias in data collections.

### 2.1 Explainable Artificial Intelligence

One of the ways of categorizing XAI methods is to divide them into local and global explanations. The local analysis aims to explain a single prediction, whereas the global one tries to explain how the whole model works in general[8].

The subcategory of **local** visual **explanations** covers such methods as attribution maps (heatmaps, saliency maps, relevance maps)[9], visualizing class-related patterns [10], or explaining by example[11]. An interesting branch of visual explanations is the category of methods based on decomposition [12]–[14] that, in contrary to optimization-based methods [15] or techniques based on sensitivity analysis, allows building self-consistent attribution maps which are consistent both in the space of models and in the input-domain [7]. For instance, Layerwise Relevance propagation (LRP)[9], [14] can be used to generate attribution maps that show parts of the image on which the classifier focused the most. Local explanations are now an actively researched topic.

On contrary, **global analyzers** are still a small part of XAI methods. Analyzing whole datasets is a tremendous task, which requires a lot of time and effort. A great manual study (manual global explanation) was presented in [16] where twelve commonly used datasets were tested. Nevertheless, some existing methods can be used to semi-automatically find repetitive errors in predictions. Semi-automatic global explanations are not only an essential tool to discover abnormalities in the whole model but in fact, this is also a tool for comparing different models and even different datasets. A common, emerging approach is to combine many local explanations into a global one. Such an approach was used to explain deep, tree-based machine learning models that are usually very hard to interpret [17]. An example of human-friendly global explanatory would be Testing with Concept Activation Vectors [18] that uses directional derivatives to quantify the importance of user-defined concepts for classification. The idea of that approach is to show natural high-level concepts, again, using local linearity. Similarly, for instance, a locally-summarized global explanation might help to create a robust adversarial example detector [19]. In the paper, we focus on one of the very first semi-automatic global explanation methods Spectral Relevance Analysis [5].

**Layer-Wise Relevance Propagation.** The general idea is to measure how pixels contribute positively and negatively to the output by decomposing the prediction function to obtain relevance scores. Hence, the goal is to attribute a contribution, or in other words, the relevance to each pixel of the image for a corresponding prediction. Bach et al. [9], propose to do that by decomposing the prediction into a sum of relevance scores for each input dimension (pixels). Those relevance scores can be visualized in a form of so-called attribution maps and show which pixels contribute positively or negatively to the output.

**Spectral Relevance Analysis** uses local explanations in the form of attribution maps for generating a summarized explanation of how the model works. The generated attribution maps are later grouped with spectral clustering, which reveals some hidden patterns forming on the attribution maps and allows the user to screen through a large dataset to find co-occurring patterns without manual, time-consuming analysis of individual explanations. The final step in this semi-supervised method is a visual inspection of interesting clusters by the user.
The steps of the method are as follows:
**Step 0.** Select batch of samples for analysis.
**Step 1.** Compute relevance scores with LRP and generate attribution maps.
**Step 2.** Normalize and preprocess the attribution maps.
**Step 3.** Perform spectral clustering on normalized attribution maps.
**Step 4.** Perform eigengap analysis to find interesting clusters.
**Step 5** (optional). Visualize selected clusters with t-SNE.
The results presented by Lapuschkin et al. [5] are very impressive, but the fact that the SPrAy method clusters the data based only on the attribution maps makes the method itself biased. This drawback makes that biased XAI focuses only on the shape of the detected objects on the attribution maps, localization of those shapes, and sometimes textures, while not considering what is under attribution maps. While the localization and shape of the attribution regions are essential, the information why the model focused on that area is even more critical. On one hand, the algorithm should take into account the colors under the attribution, the textures, and what exactly is there. On the other hand, analyzing only input images gives absolutely no insight into the inner model's workings. Hence, the main proposition of this paper is to merge both attributions and corresponding inputs. This paper proposes an improvement of the method and delivers in-depth research regarding this newly-formulated branch of global explainability methods. Details are provided in the Methodology section.

## 2.2 Bias in data

Bias in data is defined as any trend or deviation from the truth in data collection that can lead to false conclusions [20]. Bias in data might cause misinterpretation not only for highly data-dependable deep learning models but also for human experts, which makes identifying and avoiding bias in the research a long-standing topic in general [21]. Most of practical ML-related research problems start with a study on a whole population, e.g., a population of benign vs. malignant skin lesions. However, in practice, it is impossible to gather all possible cases from the whole population. The population analysis uses only a small representative group of individuals. If the sample is not well represented, conclusions will also not be generalizable[20]. For instance, if all sensitive asthma patients were carefully hospitalized during their pneumonia and hence never got any complications, the model might conclude that asthma prevents complications [22]. The influence of bias in data can be noticed in numerous applications. There is a known problem of gender and racial bias in sentiment analysis[23]. It appears that certain groups of people seem to be using specific words more often than others. When we want to analyze a slang, it could be a welcomed result, but in the case of unpolarized text, we could get a wrong prediction that was based only on the gender, race, or age of the person speaking [24]. Similarly, in the case of creditworthiness prediction in the United States, the predicted credit risks were different depending on the race [25]. Even when it comes to widely accepted by the ML community benchmark datasets, a bias still can be found. For instance, ImageNet [26] has many underrepresented classes. A car class is represented mostly by racing cars [16], and also, as reported, the ImageNet seems to be undesirably biased towards texture [27].

When it comes to skin lesion datasets [28], [29], the possible bias was already discovered in 2019 [30], but the exact source of it was not identified. The common goal of skin lesion recognition is to classify skin lesions into benign or malignant type, or to specify its exact type, to find dangerous changes early. Dermatologists support their diagnosis by careful analysis of skin lesions with a broad set of dermoscopic methods, complemented with their deep intuition. In contrast, deep models find relevant features during the training based on the provided dataset. Bissoto et al. [30] suspected that a widely used dataset of skin lesions might be biased, and hence they conducted a series of experiments regarding that matter. They used segmentation masks of each lesion and modified the dataset by covering each lesion with a black segmentation mask. The dataset modified in that way was then used to train a convolutional neural network to differentiate benign and malignant skin lesions – but without any lesions in the dataset. Surprisingly, the results showed that the model trained and tested on data without any lesions could classify them correctly with the performance (AUC) above 73%, which is only ten-percentage points less than the performance on original data. Because the shape of the skin lesion is a significant feature for dermatologists, the researchers changed segmentation masks to black boxes and repeated the experiments. The results were even more surprising because the performance was almost the same as in the previous tests. Those results raise an important question: *whether we should **blindly** trust the machine learning system based only on performance metrics*? Those metrics are always generated based on the same biased test set, which makes internal validity doubtful. However, even if we know that the bias exists, we should ask ourselves another question: ***what** exactly is the bias source and **how** to eliminate or at least mitigate it*?

Barata et al. [31] tried to find the source of bias by manual analysis of skin lesions. They concluded that the model might be sensitive to the look of a skin lesion but also black frames, skin tone, and some artifacts such as white reflections. However, manual inspection is time-consuming and may lead to overlooking some important large-scale patterns. Discovering the root of this problem is the first step to designing more robust and trustful systems. This paper attempts to answer those questions by providing a methodology that will help to find the origin of the bias in data.

## 3 Methodology

In this section, the improvement of the spectral relevance analysis is proposed, and it is shown how this method can be used for bias identification.

### 3.1 Detecting bias with GEBI

GEBI's ability to detect a few possible bias-causing artifacts is demonstrated on the example of a skin lesion dataset.

The steps of the method are as follows:

**Step 0**. Select samples for analysis.
**Step 1.** Compute attribution maps for samples of one class.
**Step 2.** Normalize and preprocess **both input samples and accompanying attribution maps** in the same manner.
**Step 3.** Reduce the dimension of each input sample and relevance map with a dimension reduction algorithm.
**Step 4.** Concatenate each reduced sample with a relevant reduced attribution map.
**Step 5.** Perform spectral clustering on reduced vectors.
**Step 6.** Visualize and analyze the obtained clusters.
**Step 7.** Formulate and test the hypothesis with the bias insertion algorithm.

Step 0 is an integral part of the analysis. Only one class should be analyzed at the same time to detect bias. Analyzing simultaneously more than a single class should be performed only in specific individual cases, e.g. when looking for possible bias-causing artifacts that could exist in each class as in the case of backdoor attacks [32].

In the first step, LRP is applied to selected input images, but any method of attribution map generation can be used.

In **Step 2**, images with contrast enhancement are normalized to bring up some clinical attributes. An additional problem here is white-balance, hence each image was preprocessed with adaptive histogram equalization.

In **Step 3**, instead of reducing the dimensionality by image-downsizing, the Isomap algorithm is applied. Direct image downsizing used in the original SpRAy method might cause loss of important small-sized features. Furthermore, most clustering algorithms have problems with handling high-dimensional data. For instance, skin lesion images look mostly similar: there is a skin lesion in the middle and (usually) lighter skin around. In medicine, very often the most interesting part are shapes and colors of detailed visible structures of skin lesions. Tiny details would disappear after the mentioned strong downsizing, whereas the general colors, similar for every lesion, would remain. In the case of nonlinear dimensionality reduction method, such as Isomap, it is possible to reduce the size nonlinearly resulting in preserving only the most important information.

The number of features should be selected individually for each kind of problem. In our case, the best results were achieved when the number of features of input images was around two times smaller than the number of attribution features. Moreover, as mentioned above, the number of features selected also depends on the chosen clustering method – many clustering methods have a problem with working on high-dimensional data.

**Step 4** is a simple concatenation of input features along with attribution features. This is a new, important step because the SpRAy method does not analyze input in conjunction with attribution maps.

It is important to note that GEBI applied standalone to the inputs also does not yield characteristic clusters as mentioned in the original SpRAy paper [5]. Those clusters seem to gather similar colored images e.g. lightly colored images are grouped together, dark ones together (see the example in Appendix). The same goes with using only attribution maps, but in contrary to inputs, here colors represent attribution. As a result, clusters are based mostly on the localization of positive/negative attributions. Unfortunately, analyzing the localization of the attribution on the attribution map standalone is not enough to find which features are important. For instance, if we had an atypical structure on the lesion on the bottom of the picture it would light up on the attribution map. This could be grouped into one cluster together with metrics, which are often at the bottom of the picture. However, in those two cases, the reason behind the attribution was different: 1) once a lesion's structure, 2) unwanted artifact (ruler). Concatenating both attention maps and inputs reduces this effect.

**Step 5** covers clustering on concatenated vectors: with features extracted from both the images and attribution maps. The difference in this step is that it is feature vectors, which are the object of clustering, and not downsized attribution maps. It is noteworthy that the user can select an arbitrary clustering method, not only spectral clustering.

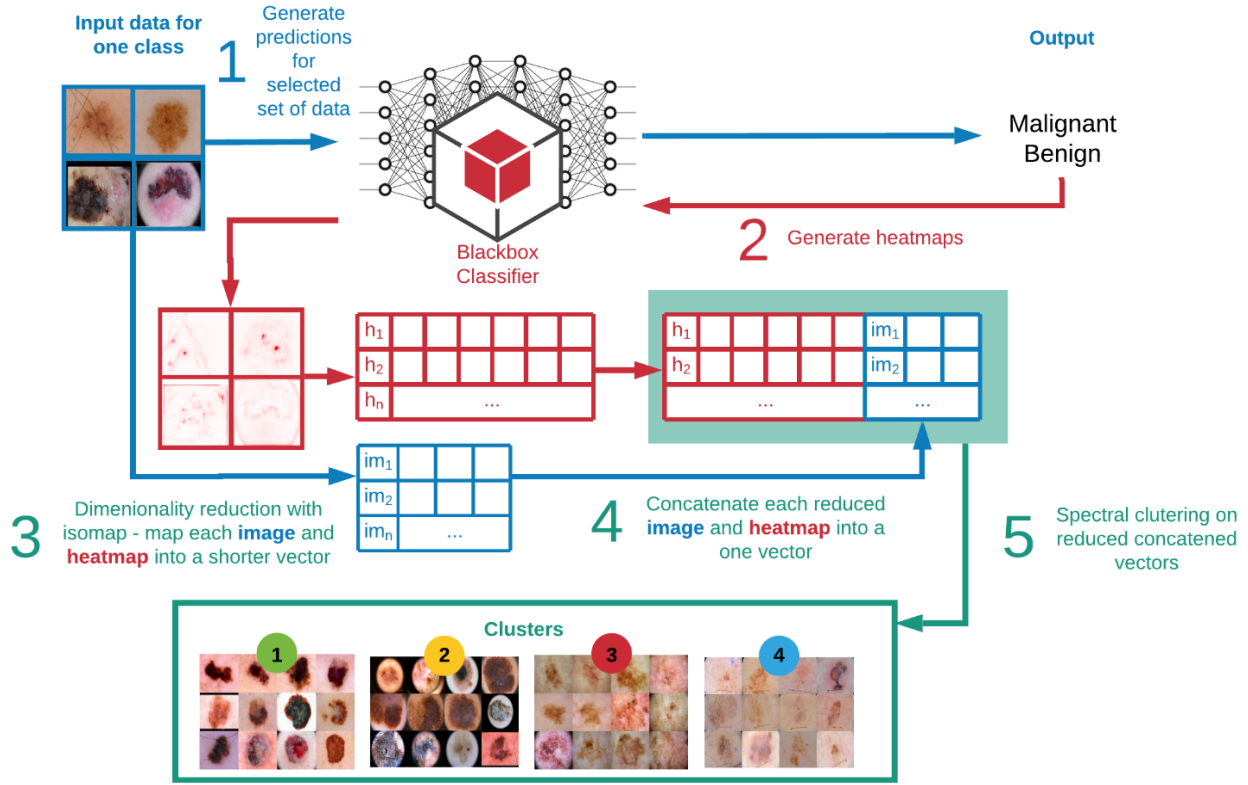

Figure 1: Pipeline of Global Explanations for Bias Identification (GEBI)

In **Step 6,** clusters are visualized in 3d-space with the Isomap algorithm. The analysis of the results of this visualization is left for the user.

Then, in the new last **step 7**, the user formulates a hypothesis about what causes the bias, for instance, the presence of artifacts in the image. The influence of the bias in data can be tested with the proposed bias insertion algorithm. The way how to test the bias is described in the next subsection. The workflow of the method is shown in Fig. 1, and the visualization of the achieved clusters in Fig. 2.

## 3.2 Bias testing – a counterfactual approach

A method to test the influence of possible bias by bias-insertion experiments is proposed. At first, like in [33], the user has to find an answer to the question: what might cause a bias? The answer can be formulated as the hypothesis and then, once the cause is identified, it should be carefully verified. For example, let us consider that in the computer vision task, in the task of for instance dog vs. cat classification, there is one cluster with dogs behind bars and no clusters of cats behind bars. Then one can think that bars might be a significant feature while classifying dogs. To test this hypothesis, we add bars to each image in the dataset and observe how the prediction score changes. If the average change of prediction is high, it means that the hypothesis is correct. Otherwise - possibly not.

The process of bias insertion is similar to different types of models and data. In the case of tabular data, for instance, in the assessment of client's creditworthiness, one can change the sex of a client and check if the model's output changes. This operation can be tested on many records and the recorded differences in prediction can be calculated and averaged afterward. Such a test would also be a crucial procedure for measuring possible unfairness.

In Natural Language Processing, in the case of sentiment analysis, we could insert bias in a similar way.

We could switch a selected word that, in our opinion, does not change the polarity of the text, to another word of the same meaning and check the change in prediction. For instance, many papers show that sentiment analyses seem to be biased by gender or race.

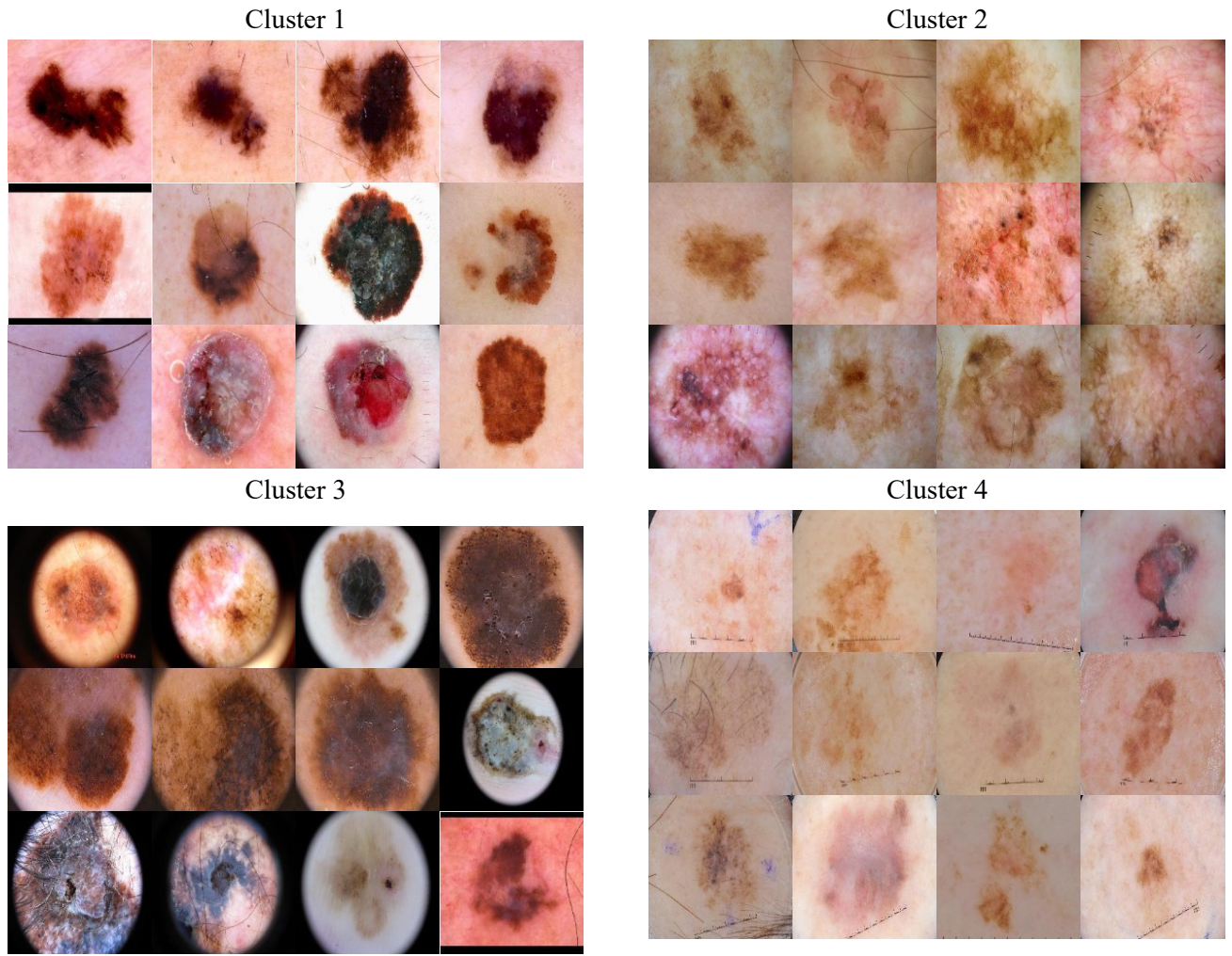

Figure 2: Sample images of four different clusters discovered with modified spectral clustering on concatenated reduced attribution maps and input images. Cluster 1 shows mostly dark skin lesions with clear border; Cluster 2 shows very textured skin lesions with numerous visible structures; Cluster 3 contains images with black frames; Cluster 4 contains mostly light-colored skin lesions with metrics, a single hair, blue markings

## 4. Experiments

In this section, we provide the information on what experiments we have conducted. Additional experiments and example results together with comparison of GEBI, SPRAY and SPRAY with Isomap reduction are delivered in the Appendix.

### 4.1 Implementation details

The training procedure presented by Mikołajczyk et al. [34] was applied, along with the widely used fine-tuned DenseNet121 [35] architecture with traditional data augmentation (rotation, zoom, shear, reflection) and early stopping. The final network had an AUC score of 0.869 on a test set.

Several types of attribution map generation were tested, including LRP, LRP flat A, LRP flat B, and Deep Taylor Decomposition (DTD). The results were similar for each type of attribution generation. The attribution maps presented in this paper were generated with DTD [36]. Each image was preprocessed with histogram equalization and contrast-enhancing. Then, the Isomap algorithm [37] was used to reduce dimensionality. Each image was reduced to the 10-dimensional vector and each attribution map to a 20-dimensional vector.

The reduced vectors were concatenated together and all vectors were clustered. The applied methods included DBSCAN, k-means, spectral clustering, affinity propagation, mean shift, OPTICS, and birch methods [38]. For the analyzed skin lesion dataset, characterized by huge intra-class variation and small interclass variation, where images seemed to be very similar, the best results were achieved with spectral clustering and traditional k-means methods. The results presented in the paper were achieved by using a spectral clustering algorithm [38]. The elbow method [39] was used to estimate the optimal number of clusters. Four clusters were found to be the most suitable solution. The clusters were examined afterward, and finally, the results were additionally visualized in a form of 3d animated plots.

## 4.3 Identification of prediction strategies

With the proposed method, four different clusters have been identified. Each cluster reveals unique characteristics in the look of the analyzed data set, which were related to skin tone, skin lesions, but also with the presence of unwanted artifacts. The first and the second cluster seem to group images based on skin lesion similarity, which is a welcome result in this case. In turn, the third cluster mostly gathers images with round or rectangular black frames, while the last, fourth cluster contains mostly light skin lesions, very often with a visible ruler. Images are presented in Fig. 2.

The proposed method is semi-automated, so a field expert should analyze the clusters. In our case, attribution was paid to clusters 3 and 4, where we identified repeating artifacts such as black frames and ruler marks. probably grouped those images. Hence, a hypothesis could be formulated that black frames and ruler marks might cause possible bias in models. To check whether those features have a significant influence on the prediction, another experiment was conducted, which consisted of inserting a possible bias and testing its influence.

## 4.4 Inserting possible bias

Since we have formulated the hypothesis that **bias in data** in the form of black frames and ruler marks **cause bias in model**, now we can examine whether it is true. To test how the prediction will change if a given feature is present in the image, model outputs were compared for the same image with and without this feature. Since removing artifacts from the images is a very complicated task, we propose to insert them instead. The goal of this step is to mimic real artifacts found in the dataset, as well as to add a new one for comparison.

**Black frames** were added to all images in the same way, without any variations in size and position. Such frames can be commonly found in numerous images, and are often recognized as unwanted artifacts [40]. Their visibility usually depends on the type of dermatoscope used.

**Ruler marks** were prepared beforehand and placed on the image in slightly different sizes, angles, and positions. Rulers are usually used by a doctor to show the size of a skin lesion on the dermoscopic image.

**Red circles** cannot be naturally found in the ISIC archive, SD-198, and Derm7pt datasets. For clear comparison, those markings have also been placed. Single red circles were placed randomly in the images, both within the skin and lesion areas. Examples of such modifications are presented in Fig. 3.

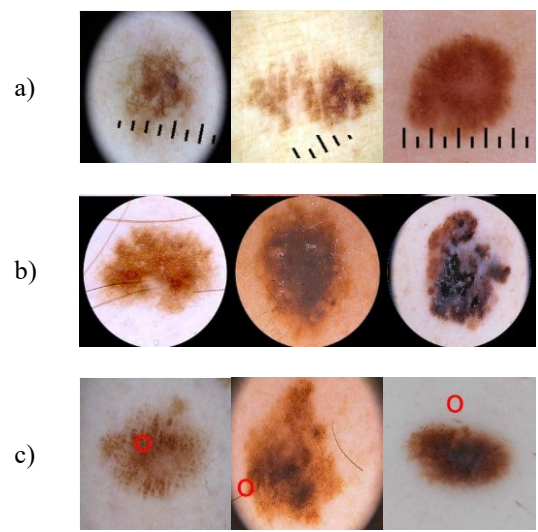

Figure 3: Samples modified by insertion of artificial bias: a) ruler markings, b) black frames, c) red circles

## 4.5 Testing bias influence

After modifying the dataset by placing selected artifacts in the images, the hypothesis was formulated as the answer to a question of whether those artifacts are causing bias in the model's performance or not. To answer this question, the effect of the presence of these artifacts, i.e. black frames, black ruler marks, and red circles, in all images on prediction changes was examined.

The idea behind testing the bias influence is illustrated in Fig. 4.

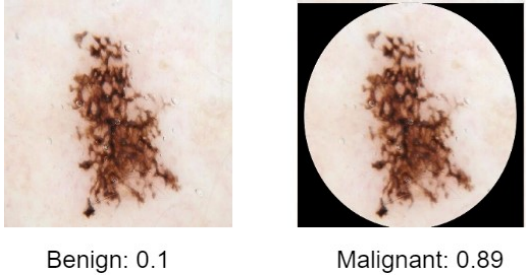

Figure 4: Idea behind counterfactual bias insertion. The model was trained to output 0 when the skin lesion is benign and 1 when it is malignant. After inserting the artifact in a form of black frame model changed the prediction score form 0.1 (benign) to 0.89 (malignant).

Differences in predictions have been calculated for 884 randomly selected malignant and benign skin lesions, separately for each type of transformation. The prediction score difference is simply a difference between the predictions on the unmodified image and after artifact insertion. Hence, the higher difference, the higher impact of the tested artifact on the final prediction. The obtained results were gathered in Table 1.

Table 1: Results in percentage points

| Added feature | Type | Average change in prediction* | Maximum change in prediction |
|---|---|---|---|
| Ruler | Mal | 2.21 | 22.01 |
|  | Ben | 1.23 | 19.91 |
| **Frame** | **Mal** | **30.77** | **62.43** |
|  | **Ben** | **32.04** | **63.66** |
| Red circle | Mal | 2.27 | 15.51 |
|  | Ben | 1.50 | 12.78 |

The highest differences in model prediction were recorded after adding a black frame to the image, whereas the introduction of a ruler and red circle did not change prediction scores much on average. An interesting part of this experiment was that the black frame did not change in any way how the skin lesion looked like, but at the same time prediction changes were very high for both malignant and benign skin lesions. On average, every output changed by 33%. Moreover, adding this type of artifact seemed to bias the model toward classifying a skin lesion as a malignant. The number of images classified as malignant raised from 31 to 228 when tested on the benign dataset. Hence, 197 out of 884 skin lesions switched prediction to malignant, considering the classification threshold equal to 0.5. This means that about 22.29% of the checked skin lesion samples changed their classes after introducing such slight modification. Black frames usually do not cover any part of skin lesion, hence such a significant change in prediction score should wake up some doubts in models' behavior. It is a very interesting finding, which should be taken into consideration while training new models in the future.

Ruler marks caused, on average, only a slight difference in model predictions, of about 1.23 and 2.21 pp., but still, it might be a dangerous reaction in some cases when the change in prediction is high. What is interesting, for those markings, there were a few cases that changed model's decision from malignant to benign in both subsets.

Adding a red circle did not make a huge difference in the output, but surprisingly, it was quite similar to the average change for ruler placement. A small number of approximately 1.5% of images switched prediction from benign to malignant. A possible reason for this is that part of malignant skin lesions tends to have atypical structures: blobs, dots, or streaks [41]. The red circle might be similar in some way to dermatological attributes. Those structures are defined, for example, in the 7-check point list or in the ABCD rule [41].

### 4.6 Code and data availability

The developed source code, user-friendly tutorials, and generated attribution maps for quick experiments with GEBI are available at github.com/agamiko/gebi. Additionally, we present source codes for SpRAy and adding bias such as black frames and ruler. The source code for LRP is available at github.com/albermax/innvestigate. The source code for clustering and Isomap reduction is available at scikitlearn. The dataset of skin lesions is available at isic-archive.com.

## 5 Discussion

Currently, the subject of interpretable and explainable artificial intelligence is constantly rising. More and more people are aware that machine learning (ML) and deep learning models require extensive testing and that their inner work should be known. Unfortunately, bias in data is still not widely discussed.

Authors of real-data applications usually test their models only in terms of accuracy performance or computation efficiency. That approach to production ML should be changed, especially when tackling safety-critical systems.

The paper presents a new method that can be used for detection of bias in data collection, or in model's behavior. The problem of biased XAI is introduced which might lead to incorrect interpretation by the model's decision-making process. The obtained results are illustrated on the example of skin lesion classification task. After a few simple but effective modifications of the SpRAy, the new GEBI method gained a significant improvement. For example, it allowed detecting that black frames, commonly existing in skin lesion dataset images, have a significant impact on model predictions. The hypothesis regarding bias in skin lesion dataset has been tested with the developed bias insertion algorithm. In fact, each image was predicted with about 32 percentage points more towards malignant skin lesions when added a black frame, which confirmed the suspicions of many researchers from the past[42]–[44].

However, bias detection and confirmation is just the first step of making models more reliable and robust. The next step should be further development of this approach. Improvement can be reached e.g. by deleting bias from datasets. Many researchers have tried to remove artifacts as the first preprocessing step before [42]–[44], although removing all of the biases is nearly impossible and does not solve the problem. Another possible approach is making the model focus on the right features. This can be done with special data augmentation. For instance in speech recognition, it can be done by randomly removing low-energy parts of the recording[45]. In our case, it could be done by randomly inserting bias into images during the training, similar to online data augmentation.

And finally, a model can be forced to focus on important parts of data, for example by attribution-training [31]. Such an approach modifies the loss function to check not only the model classification performance but also whether it focuses on the right regions.

The results are presented in an open-science manner, and relevant codes for both the proposed method and bias insertion are provided.

# 6 Acknowledgments

The research reported in this publication was supported by Polish National Science Centre (Grant Preludium No: UMO-2019/35/N/ST6/04052). The authors wish to express their thanks for the support.

# Appendix

## Experiments

We present results on the original SpRAy, on SpRAy used with Isomap reduction, and with GEBI (input and attribution map together with Isomap). For repeatability of all experiments we provide the same random seed, preprocessing methods, and the same number of clusters n = 4. The number of clusters was assumed based on conducted experiments.

For every experiment, we present both attribution maps and images on the 3D visualization. Each color represents a different cluster. Provided visualization helps to understand how clusters changes depending on what data was used as input: attribution maps, images, or both, and also with/without Isomap. Additionally, to keep the clarity of figures, for every experiment, we show just 15 first images from each cluster (due to alphabetical order).

### 1.1 Clustering based only on heatmaps with image resize – original SPRAY

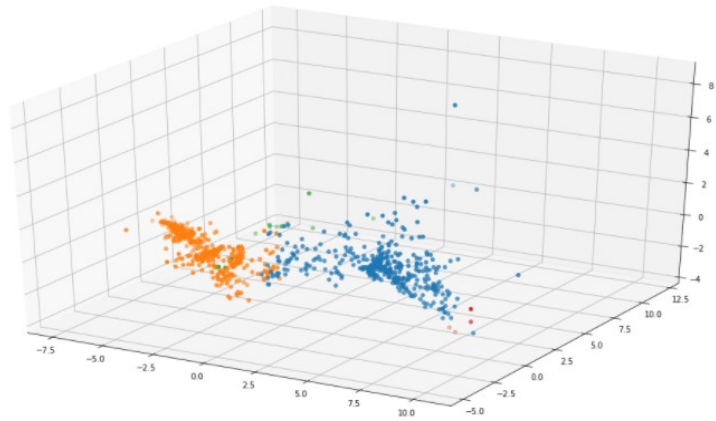

Figure 1: Attribution maps presented on 3D space – original Spray on attribution maps

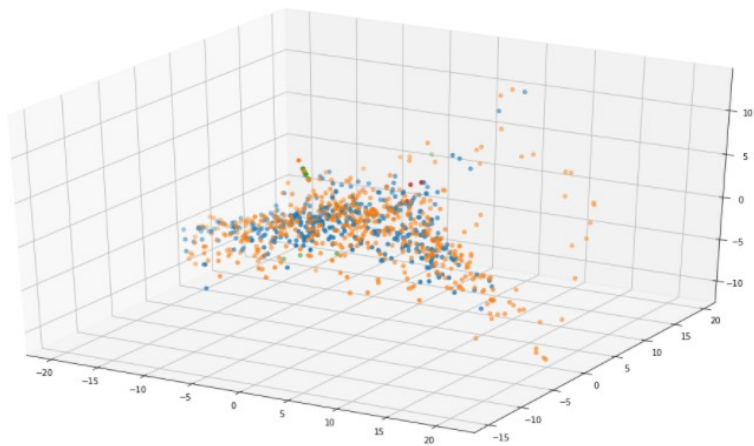

Figure 2: Images presented on 3D space - original Spray only on attribution maps

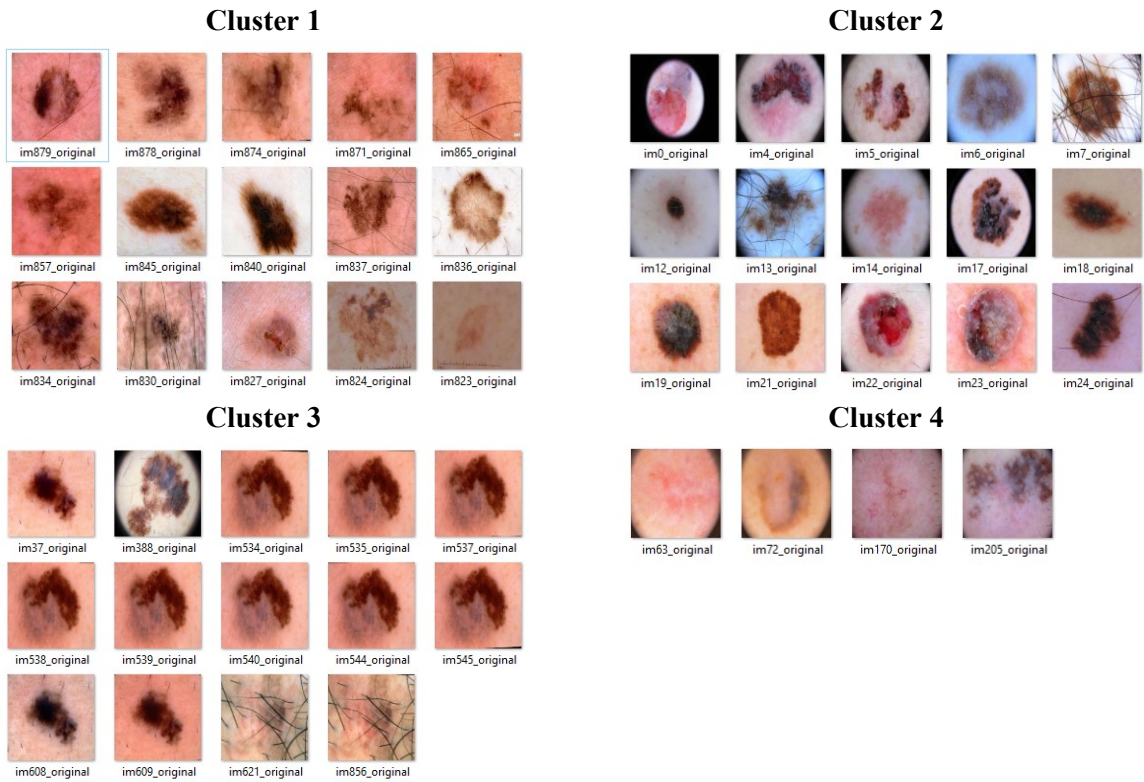

Figure 3: Resulting clusters (Spray on attribution maps) - 15 first images from each cluster due to alphabetical order

**Comment**: We can see that in the attribution visualization two main clusters emerge. The other two clusters are smaller and contain respectively 14 and 4 samples. The clustering algorithm takes into account only attribution maps, hence as shown in figure 2 images are not well separated.

## 1.2 Clustering based only on images with image resize – original SPRAY

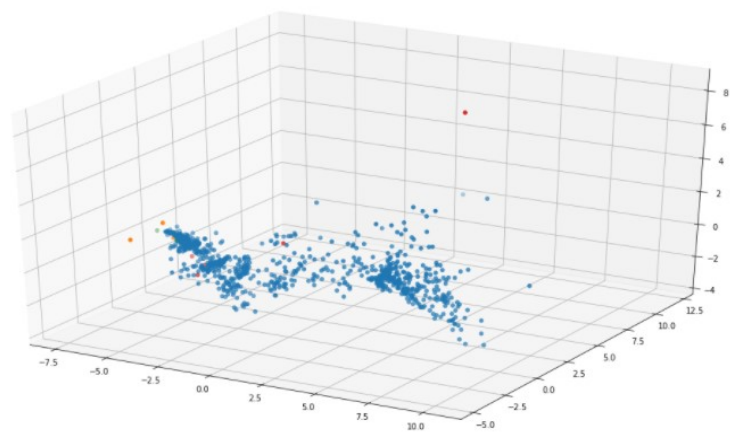

Figure 4: Attribution maps presented on 3D space – original Spray on input images

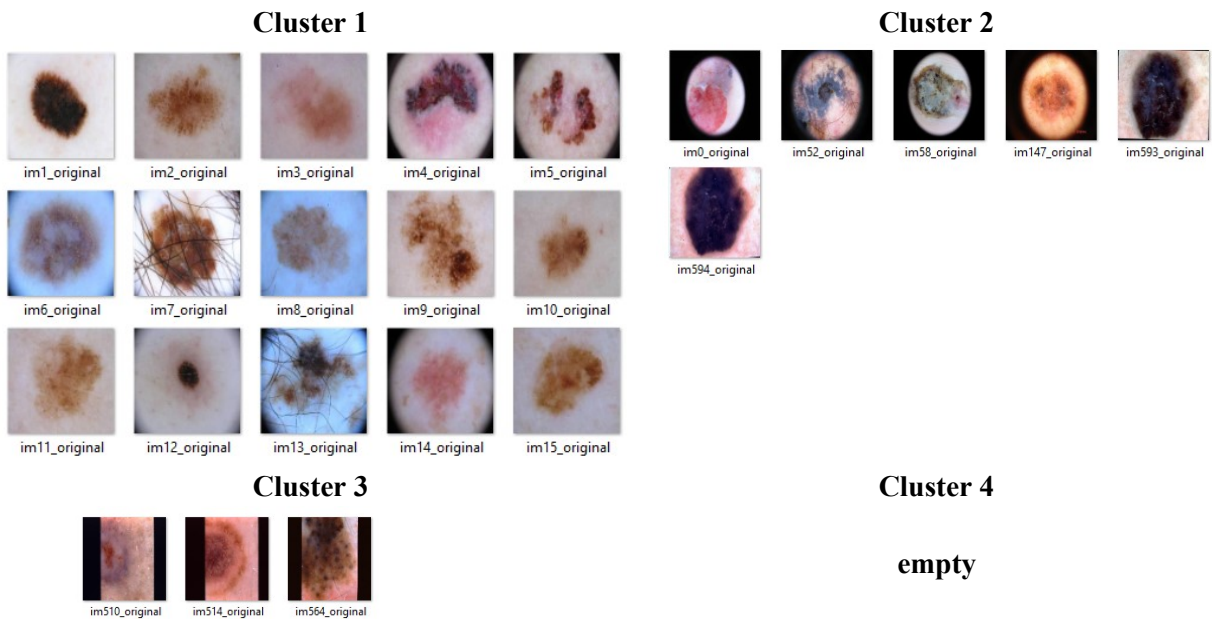

Figure 6: Images presented on 3D space – original Spray on input images

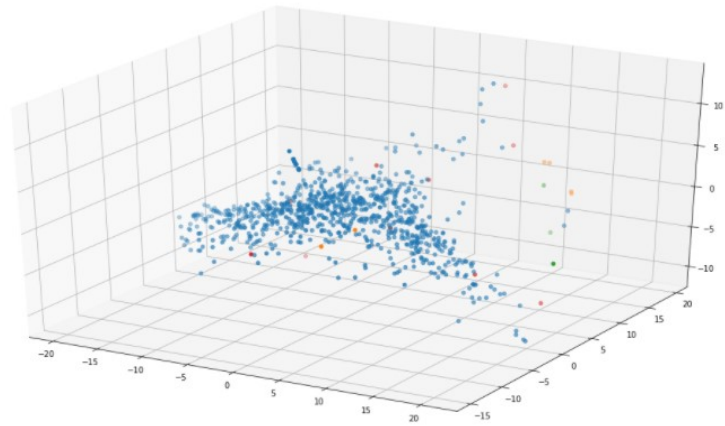

Figure 5: Resulting clusters (Spray on input images) - 15 first images from each cluster due to alphabetical order

**Comment**: We can see (figure 4) that in the attribution visualization one main cluster emerges. The other two clusters are smaller and contain respectively 10 and 1 sample. The last cluster remains empty. The clustering algorithm takes into account only input images, which downsized look very similar. As shown in figure 5 and 6 images are not well separated and grouped mostly into one cluster.

## 1.3 Clustering based only on heatmaps – SPRAY modified with Isomap dimension reduction

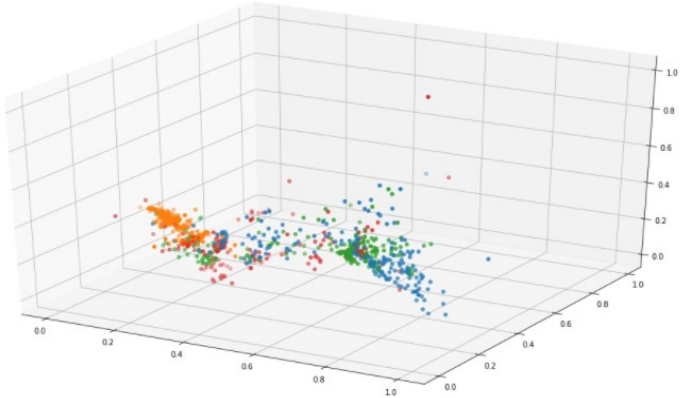

Figure 7: Attribution maps presented on 3D space – modified Spray (with Isomap) on attribution maps

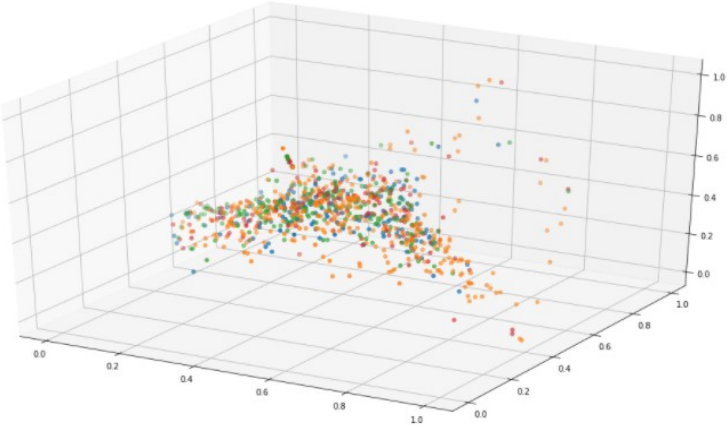

Figure 8: Inputs presented on 3D space – modified Spray (with Isomap) on attribution maps

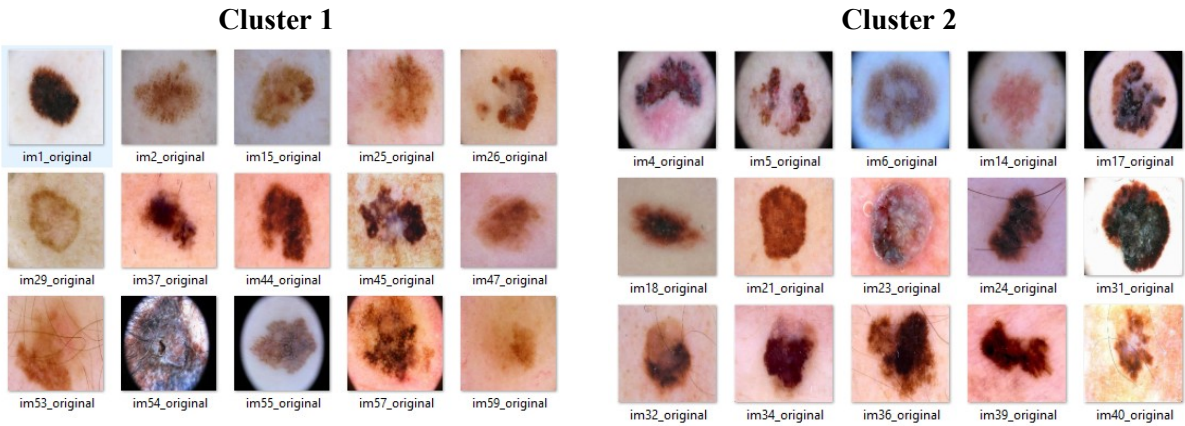

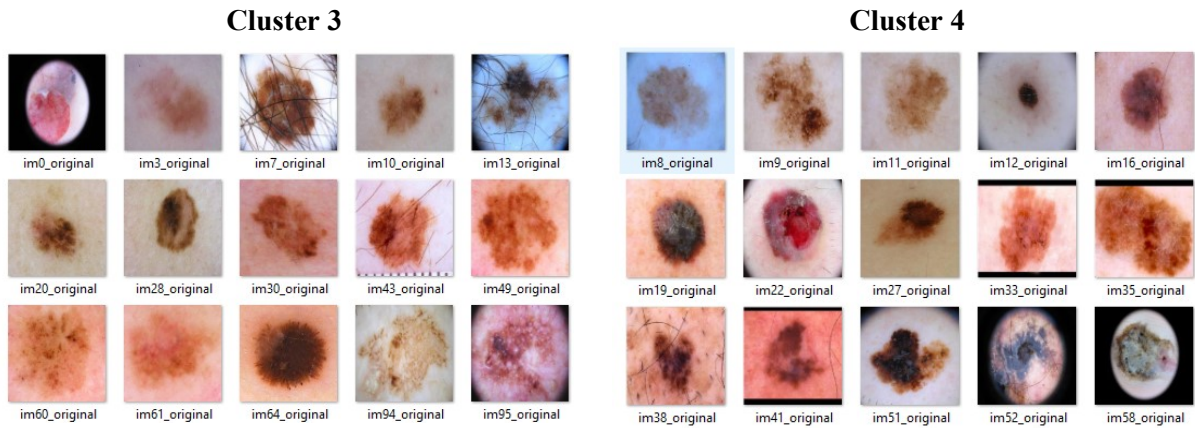

Figure 9: Resulting clusters (Spray on attribution maps with Isomap reduction) - 15 first images from each cluster due to alphabetical order

**Comment**: We can see (figure 7) that in the attribution visualization four different clusters emerge. All clusters have similar sizes. The clustering algorithm takes into account only attribution maps, hence as shown in figure 8 images are still not well separated. For example, images with black frames can be found in all clusters.

### 1.4 Clustering based only on input images – SPRAY modified with Isomap dimension reduction

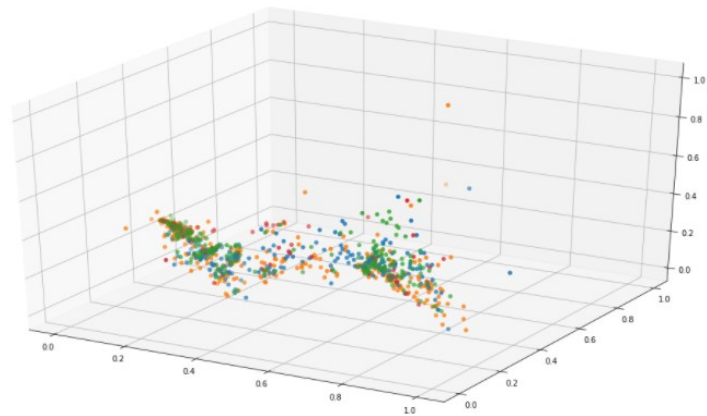

Figure 10: Attribution maps presented on 3D space – modified Spray (with Isomap) on input images

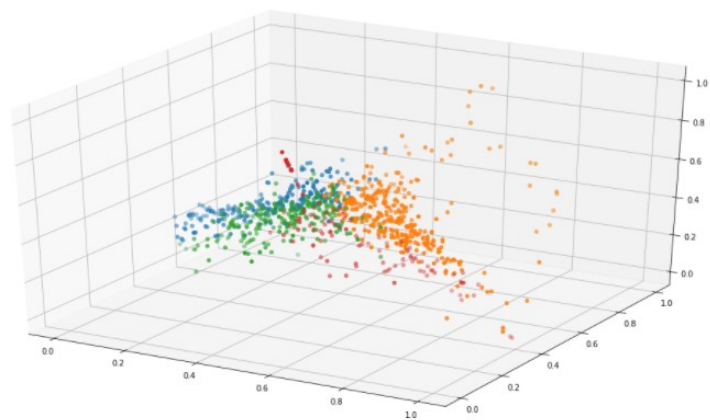

Figure 11: Inputs presented on 3D space – modified Spray (with Isomap) on input images

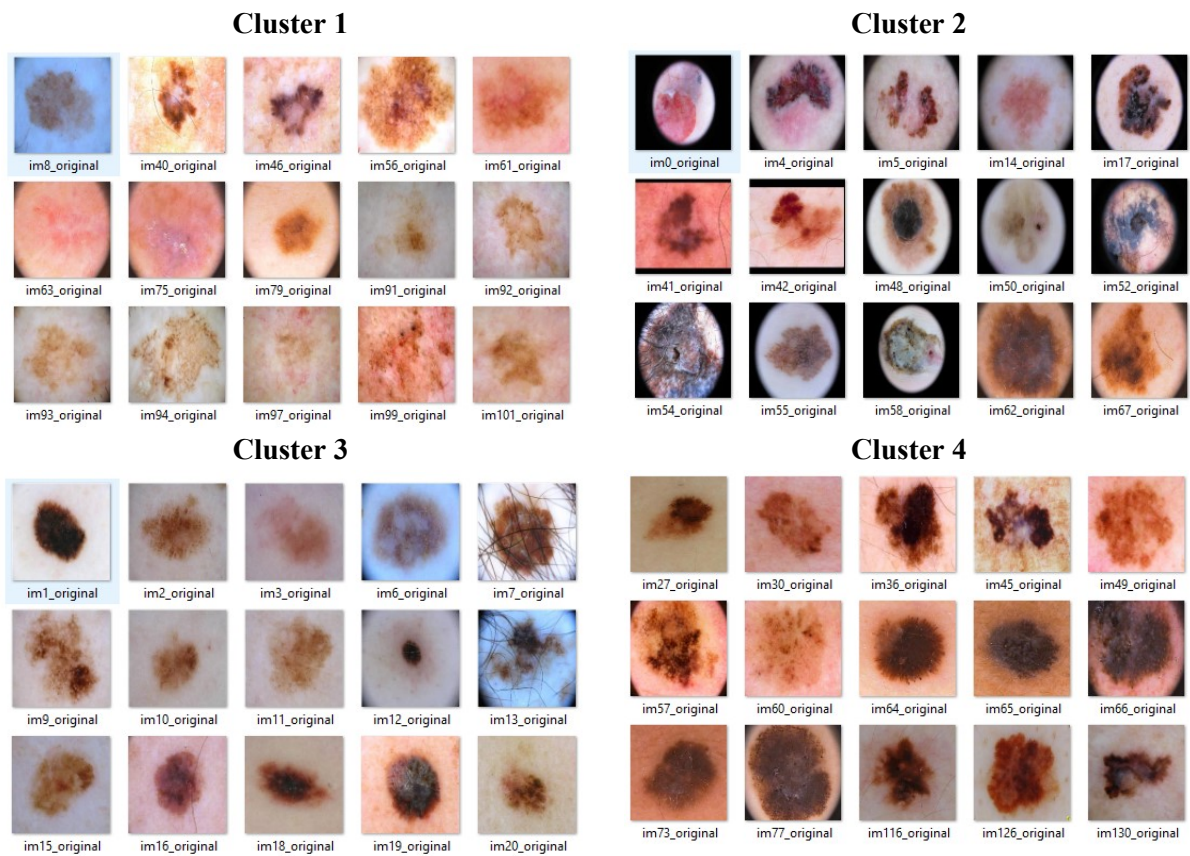

Figure 122: Resulting clusters (Spray on input images with Isomap reduction) - 15 first images from each cluster due to alphabetical order

**Comment**: We can see (figure 10) that in the attribution visualization, four different clusters emerge but attribution maps are not well separated. On the other hand, this time images (figure 11) are much better separated. It is clearly visible that clustering is based mostly on the color of images, in this case (figure 12).

## 1.5 Clustering based on heatmaps and input images with Isomap dimension reduction - GEBI

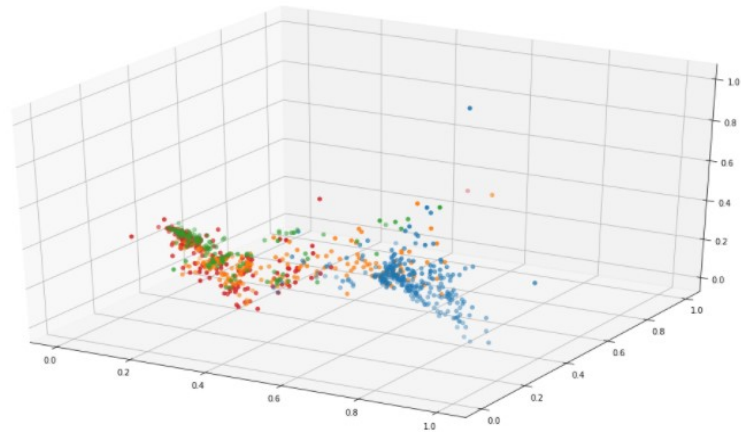

Figure 13: Attribution maps presented on 3D space – GEBI

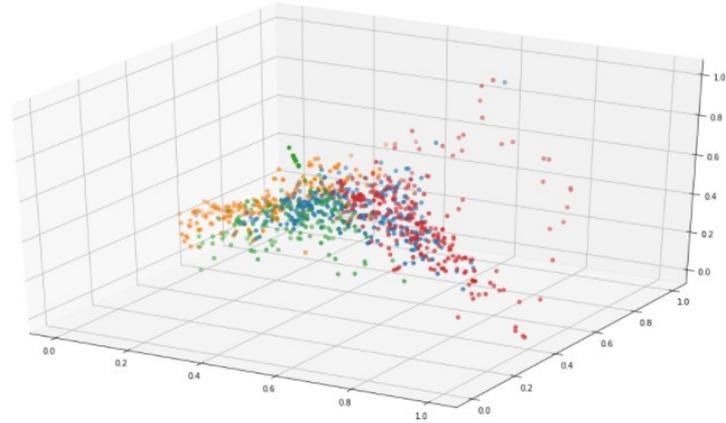

Figure 14: Inputs presented on 3D space – GEBI

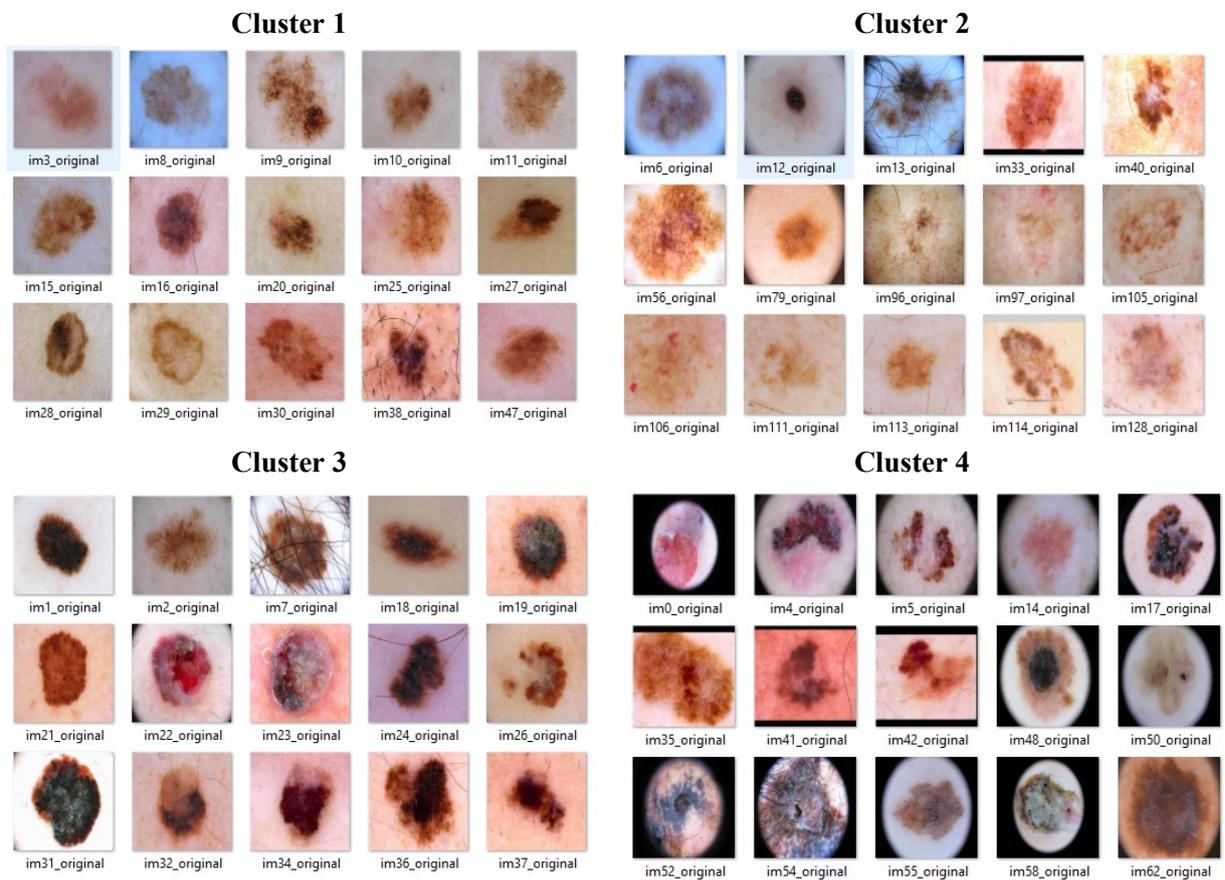

Figure 15: Resulting clusters (GEBI) - 15 first images from each cluster due to alphabetical order

**Comment**: Clustering jointly on both attribution maps and images resulted in the different results of clustering than analyzing images or attribution maps alone. In contrary to clustering only heatmaps, we can easier evaluate and analyze the results. Moreover, the clustering is not as biased towards the color and white balance of the images, as in the case of clustering only input images. For example, cluster 4 shows images with black frames whereas cluster 2 catches most of the lesions with ruler marks mentioned in the paper.

# Source Code

We share source code on GitHub repository (github.com/AgaMiko/GEBI) to enable the readers for conducting additional experiments i.e. testing different clustering algorithms, evaluating a different number of clusters, or other parameters.